\documentclass[letterpaper, 10 pt, conference]{ieeeconf}
  \twocolumn                                                        
\IEEEoverridecommandlockouts                              

\usepackage{dblfloatfix}
\usepackage{epsfig}
\usepackage{psfrag,graphicx,color}
\usepackage[normalem]{ulem}
\usepackage{amsmath,amsfonts}
\usepackage{hyperref}
\usepackage{epsfig,psfrag,graphicx,color}
\usepackage{amsthm}
\usepackage{amssymb}
\usepackage{algpseudocode}
\usepackage{algorithm}
\usepackage{color} 
\usepackage{url} 
\usepackage{cleveref}
\usepackage{tabularx}
\font\bfmath=cmmib10
\mathchardef\Gamma="7100
\mathchardef\Delta="7101
\mathchardef\Theta="7102
\mathchardef\Lambda="7103
\mathchardef\Xi="7104
\mathchardef\Pi="7105
\mathchardef\Sigma="7106
\mathchardef\Upsilon="7107
\mathchardef\Phi="7108
\mathchardef\Psi="7109
\mathchardef\Omega="710A

\mathchardef\alpha="710B
\mathchardef\beta="710C
\mathchardef\gamma="710D
\mathchardef\delta="710E
\mathchardef\epsilon="710F
\mathchardef\zeta="7110
\mathchardef\eta="7111
\mathchardef\theta="7112
\mathchardef\iota="7113
\mathchardef\kappa="7114
\mathchardef\lambda="7115
\mathchardef\mu="7116
\mathchardef\nu="7117
\mathchardef\xi="7118
\mathchardef\pi="7119
\mathchardef\rho="711A
\mathchardef\sigma="711B
\mathchardef\tau="711C
\mathchardef\upsilon="711D
\mathchardef\phi="711E
\mathchardef\chi="711F
\mathchardef\psi="7120
\mathchardef\omega="7121
\mathchardef\epsilon="7122

\mathchardef\varepsilon="7122
\mathchardef\vartheta="7123
\mathchardef\varpi="7124
\mathchardef\varrho="7125
\mathchardef\varsigma="7126
\mathchardef\varphi="7127
\mathchardef\imath="717B
\mathchardef\jmath="717C

\def\bfp{{\mbox{\boldmath $p$}}}

\def\smallbfW{{\raise1.5pt\hbox{\mbox{\boldmath $_W$}}}}

\def\Re{\mathbb{R}}

\def\mypsfrag#1#2#3#4#5{
        \begin{figure}[htp]
           \begin{center}
              {\leavevmode
                 {\includegraphics[width=#1truecm]{#2.eps}}
              }
           \end{center}
           \vspace{#3}
           \caption{#4}
           \label{#5}
        \end{figure}
}

\def\my4psfrag#1#2#3#4#5#6#7#8{
        \begin{figure}[htp]
        \begin{center}
            \begin{tabular}[h]{c c}
              {\leavevmode{\includegraphics[width=#1truecm]{#2.eps}}}
              &
              {\leavevmode{\includegraphics[width=#1truecm]{#3.eps}}} \\
              {\leavevmode{\includegraphics[width=#1truecm]{#4.eps}}}
              &
              {\leavevmode{\includegraphics[width=#1truecm]{#5.eps}}}
         \end{tabular}
           \vspace{#6}
           \caption{#7}
           \label{#8}
        \end{center}
        \end{figure}
}

\def\mydouble4psfrag#1#2#3#4#5#6#7#8{
        \begin{figure*}[htp]
        \begin{center}
            \begin{tabular}[h]{c c}
              {\leavevmode{\includegraphics[width=#1truecm]{#2.eps}}}
              &
              {\leavevmode{\includegraphics[width=#1truecm]{#3.eps}}} \\
              {\leavevmode{\includegraphics[width=#1truecm]{#4.eps}}}
              &
              {\leavevmode{\includegraphics[width=#1truecm]{#5.eps}}}
         \end{tabular}
           \vspace{#6}
           \caption{#7}
           \label{#8}
        \end{center}
        \end{figure*}
}

\def\bfp{{\mbox{\boldmath $p$}}}

\usepackage{amssymb}  
\usepackage[table,xcdraw]{xcolor}

\def\va#1#2{X_{#1,#2}}

\Crefname{equation}{Eq.}{Eqs.}
\Crefname{figure}{Fig.}{Figs.}
\Crefname{tabular}{Tab.}{Tabs.}
\Crefname{table}{Tab.}{Tabs.}
\Crefname{section}{Sec.}{Secs.}

\newcommand{\footref}[1]{%
    $^{\ref{#1}}$%
}

\newlength{\Oldarrayrulewidth}

\usepackage{hhline}

\renewcommand\thesubsubsection{T\arabic{subsubsection}}

\makeatletter

\makeatletter
\IEEEtriggercmd{\reset@font\normalfont\fontsize{7.5pt}{8.40pt}\selectfont}
\makeatother
\IEEEtriggeratref{1}

\begin{document}

\title{\LARGE \bf  
A Mixed-Integer Linear Programming Formulation for Human Multi-Robot Task Allocation  
}

\author{ Martina Lippi, Alessandro Marino 
\thanks{${}^1$Roma Tre University, Italy, {\tt\small martina.lippi@uniroma3.it} }
\thanks{{${}^2$University of Cassino and Southern Lazio, Italy
	{\tt\small al.marino@unicas.it}}}
  }
\maketitle


\begin{abstract}
    In this work, we address a task allocation problem for human multi-robot settings. Given a set of tasks to perform, we formulate a general Mixed-Integer Linear Programming (MILP) problem aiming at minimizing the overall execution time while optimizing the quality of the executed tasks as well as human and robotic workload. Different skills of the agents, both  human and robotic,  are taken into account and human operators are enabled to  either directly execute tasks or play  supervisory roles; moreover, multiple manipulators can tightly collaborate if required to carry out a task.  Finally, as realistic in human contexts,  human parameters are assumed to vary over time, e.g., due to increasing human level of fatigue. Therefore, online monitoring is required and re-allocation is performed if needed. Simulations in a realistic scenario with two manipulators and a human operator performing an assembly task validate the effectiveness of the approach. 
\end{abstract}

\section{Introduction }
In the last decade, Human–Robot collaboration (HRC) has received increasing interest
of the research community, as well as of the
the industrial world.
The benefits of  having humans and robots cooperating in the execution of complex tasks 
 lay in the possibility  of achieving flexible and highly reconfigurable production systems and of easily changing 
 task execution  to accommodate different
product families while meeting high quality standards~\cite{Tsarouchi_2016}.

Antithetical yet complementary capabilities characterize human and robot entities: reasoning and manipulation skills for the human component, strength and endurance capabilities for the robotic one. The different abilities make humans and robots better  suited to different sets of tasks, e.g.,  the former are more apt to handle objects of small size or with particular shapes and/or materials, while the latter are more appropriate for manipulating heavy objects with regular shapes or for performing repetitive tasks~\cite{Hashemi_ARC2020}.

Indeed, a core component of HRC is how to optimally distribute tasks  between robotic and human counterparts; despite the potentialities of human-robot collaborative scenarios, task allocation in these setups is far from trivial since there are a variety of tasks that can be performed by both and for which  it must be decided to whom to assign execution while maximizing certain criteria. Furthermore, several tasks also exist that benefit from the simultaneous collaboration of humans and robots, e.g., robots holding large parts for assembly and humans performing operations on these parts~\cite{Elzatari_RAS2019}. Concerning the optimization criteria, different factors should be taken into account such as overall completion time,  human fatigue, and specific expertise. In addition, human parameters are generally challenging to be quantified and can vary over time.

In this work, we propose a novel flexible task allocation formulation in which multiple manipulators and human operators are involved. In the proposed framework,  each agent, either human or robotic, can be differently skilled at a given task. Multiple manipulators can tightly collaborate if required to carry out a task, while human operators can either directly execute tasks or play a \emph{supervisory} role, allowing to guarantee a minimum task execution accuracy.  Indeed, the supervisory role can be useful, for example,  in case where the robotic component is not fully capable at a task and the person monitors its activity to prevent errors from occurring. Moreover, precedence as well as spatial constraints are taken into account by the devised framework with the latter taking into consideration that, due to workspace limitations, some tasks might not be carried on simultaneously.  

The devised solution aims at minimizing the overall required time while optimizing the quality of the executed tasks as well as human and robotic workload. 
Moreover, as realistic in human contexts, we consider that human parameters can vary over time.  Therefore, the proposed framework also comprises online monitoring and possible 
 re-allocation whenever required to meet future constraints as well as to preserve the solution optimality to a certain extent.

\section{Related work}\label{sec:relwork}
Task allocation and scheduling for multi-robot  as well for human-robot teams are important problems with applications to manufacturing, warehouse automation, and pickup-and-delivery~\cite{Nunes_RAS2017}.
Human-Robot and Multi-Robot Task Allocation (HRTA and MRTA, respectively) share many common points and methodologies but also have inherent differences due to peculiarities of human agents.
Indeed, the problem of MRTA is formally addressed in \cite{Mataric_2004} by combining operations research and combinatorial optimization,  and how this formalism can be used for the synthesis of new approaches is shown.
Since then, several approaches have been proposed addressing many aspects and problems concerning both MRTA and HRTA as, for example, distributed implementation; this is the case in~\cite{Suslova_IROS2020} where authors focus on a distributed multi-robot setting with tasks having time window and ordering constraints. 

The work in~\cite{gombolaytercio_2013rss} formulates a MILP problem in which temporal and spatial constraints are taken into account in the framework of HRTA and re-planning is  envisaged as well. In detail, the solution is a multi-agent task sequencer that is inspired by real-time processor scheduling techniques and is adapted
to leverage hierarchical problem structure. The same authors  in \cite{Gombolay_2015} conduct experiments to establish the effect on the task allocation solution of the team composition and of the level of  influence authority. In this perspective, the allocation could be  completely made by human,  semi-autonomous, with the human deciding which task to allocate to himself/herself, or autonomous, i.e., the robot allocates all tasks.
The authors  found that an autonomous robot can outperform a human worker in the execution of part or all of the process of task allocation, and that people preferred to cede their control authority to the robot. 

A closed-loop planning framework which adapts  to varying human parameters and task sets is presented in \cite{Shannon_ACC2016}.  Highly coupled tasks between humans and robots are considered  and  human parameters like task execution time, workload and performance   are adopted to update in real-time  parameters of human model with the purpose or replanning whenever required.

The work in~\cite{Bogner_2018}  considers a scenario requiring  processing multiple parts, each consisting of a series of tasks constrained by  precedence relationships between them. Tasks are executed by
human workers and robots which are resources required for processing tasks, and the problem is identified as a simplification of the Multi-Mode-Resource-Constrained-Project-Scheduling problem. Differently from previous works, additional constraints, such as minimum distance requirements and waiting times for agents, are added in order to enhance the fluency of the mixed human-robot team.
Finally, heuristics are proposed by the authors to solve the problem and mitigate the computational burden required by the proposed model.

The problem of task allocation in a framework where  single human is involved  is presented in \cite{Zhang_ICRA2020}. The assembly scheduling problem is adaptive in the sense that it is modulated according to human capabilities: the latter is constantly monitored and re-allocation is carried out if its variation is above a pre-defined threshold; a genetic algorithm is then used to solve the optimization problem. 

In \cite{Malik2019ComplexitybasedTA}, the task distribution between humans and robots is addressed by considering a quantitative classification of task complexity. The latter considers physical features of components to be assembled and tasks are differentiated in high-complexity tasks due to the inherent complexity of handling and mounting components and in low-complexity tasks. It is shown that such a classification lowers deployment and changeover times.

Human factors are considered also in~\cite{Makrini_2019}, where task allocation in human-robot assembly applications is addressed  by considering human capabilities and workload as well as  ergonomics aspects based on the human body posture.

Similarly to previous work, in~\cite{Lamon_RAL2019} factors to be considered in allocating
tasks among a set of agents made of humans and robots in industrial manufacturing scenarios are considered.
Tasks are decomposed in atomic actions and are  offline allocated  based on indexes like complexity, agent dexterity, required effort, and experiments are carried out to evaluate system's performance.

Finally, work in~\cite{Wang_RAL2020} focuses on large-scale instances in which algorithmic solutions can be not suitable and proposes a graph attention network-based scheduler which leverages imitation learning to learn from expert demonstrations on small scale problems. 

With respect to previous works, the following novelties are introduced:
\begin{itemize}
\item the offline task allocation takes into account several features and constraints simultaneously, like time and spatial constraints, and it allows to assign more than one agent to the same task; moreover, the concept of human supervision is introduced in order to increase  the  quality of task execution;
\item  a closed-loop re-scheduling framework is devised that, based on real-time monitoring of parameters performance, is able to 
adapt to dynamic and stochastic agent parameters.

\end{itemize}

\section{Problem setting}
\begin{table}[h!]
\centering
\caption{Main notations introduced in the paper. 
}
\begin{tabularx}{\linewidth}{|l|X|}
\hline
Variable & Meaning  \\ \hline
    $m$        & Number of tasks  \\ \hline
    $n_a$        & Number of agents  \\\hline
    $n_r$        & Number of robot agents \\ \hline
    $n_h$        & Number of human agents  \\ \hline
    $M\in\Re$ & Arbitrary large positive constant  \\ \hline
    $\mathcal{T}$        & Set of tasks $\mathcal{T}=\{\tau_1, \tau_2, \cdots, \tau_m\}$   \\ \hline
    $\mathcal{A}$        & Set of agents $\mathcal{A}=\{a_1, a_2, \cdots, a_{n_a}\}$ comprising humans and robots  \\ \hline
    $\mathcal{A}_r$        & Set of robot agents $\mathcal{A}_r=\{a_{r,1}, a_{r,2}, \cdots, a_{r,n_r}\}$   \\ \hline
 $\mathcal{A}_h$        & Set of human agents $\mathcal{A}_h=\{a_{h,1}, a_{h,2}, \cdots, a_{h,n_h}\}$   \\ \hline
 $g(\tau_i)$   &    Task category  to which $\tau_i$ is associated    \\ \hline
$\va{i}{j}\in\{0,1\}$        & Binary decision variable for the assignment of task $\tau_i$ to agent $a_j$  \\\hline
$S_{i,j}\in\{0,1\}$   & Binary decision variable for the assignment of human supervision of task $\tau_i\in\mathcal{T}$ to human agent $a_{h,j}\in\mathcal{A}_h$ 
 \\ \hline    
    $P_{i,k}\in\{0,1\}$ & Binary variable denoting a precedence constraint, i.e., if $P_{i,k}=1$ task $\tau_i$ must end before task $\tau_k$, otherwise no order is assigned  \\ \hline
      $D_{i,k}\in\{0,1\}$ & Binary variable denoting if there is a spatial constraint, i.e., if $D_{i,k}=1$ task $\tau_i$ and $\tau_k$ require to occupy the same location. \\ \hline
    $C_{i}\in\{0,1\}$   & Binary variable which is $1$ if task $\tau_i$ requires two agents to be executed (collaborative task), $0$ otherwise   \\ \hline
    $\underline{t}_{i},\overline{t}_{i}$        & Start and end times of task $\tau_i$  \\\hline
    $\Delta_{i,j}$  &  Execution time required by agent $a_j$ to perform task $\tau_i$ 
     \\\hline
    $T_M$  & Maximum overall execution time 
     \\\hline
    $q_{i,j}$ &  Execution  quality index for executing task $\tau_i$ by agent $a_j$ 
     \\ \hline  
    $q^s_{i,j}$ &  Supervision quality provided by human agent $a_j\in\mathcal{A}_h$ for task $\tau_i\in\mathcal{T}$ \\ \hline     
    $w_{i,j}$& Workload of agent  $a_j$ in performing task $\tau_i$      \\ \hline
    $w^s_{i,j}$& Workload of human agent  $a_{h,j}$ in supervising task $\tau_i$ 
    \\ \hline
    $V_{i,k,j}\in \{0,\, 1\}$& Auxiliary variable for allocating at most one task to each agent at each time  \\ \hline
      $Z_{i,k}\in \{0,\, 1\}$ &  Auxiliary variable for spatial constraints  \\ \hline
      $\delta$ & Re-allocation parameter \\ \hline
\end{tabularx}%
\label{tab:notation}
\end{table}

In this section, we describe the problem setting and introduce the main variables of the paper, which are summarized in 
Table \ref{tab:notation}. 
We consider a human-multi-robot collaborative  scenario in which  there are several tasks to accomplish, with different requirements and constraints, and there are several agents to which these tasks might be allocated. 
In detail, agents are divided into human and robot agents. 
We denote   the set of agents by $\mathcal{A}=\{a_1, a_2, \cdots, a_{n_a}\}=\mathcal{A}_h\cup \mathcal{A}_r$, with  $\mathcal{A}_h=\{a_{h,1}, a_{h,2}, \cdots, a_{h,n_h}\}$   the set of human agents with cardinality $n_h$,  $\mathcal{A}_r=\{a_{r,1}, a_{r,2}, \cdots, a_{r,n_r}\}$ the set of robotic agents with cardinality $n_r$ and $n_a=n_h+n_r$. A set  $\mathcal{T}=\{\tau_1, \tau_2, \cdots, \tau_m\}$  of $m$ tasks with cardinality $m$ is defined in which: 
\begin{itemize}
    \item precedence constraints are defined, i.e., a binary variable $P_{i,k}$ is introduced for each couple of tasks $\tau_i,\tau_k$ and is $1$ if task $\tau_i$ must end before task $\tau_k$, $0$ otherwise.  This allows to express sequentiality of tasks;
	\item estimated execution times by each agent are defined, i.e., $\Delta_{i,j}$ is the execution time required by agent $a_j$ to perform task $\tau_i$; 
	\item estimated workloads for each agent are defined, i.e., we denote the workload for agent $a_j$ to execute  task $\tau_i$ by~$w_{i,j}$; 
    \item spatial locations are assigned, i.e., for each task $\tau_i$ the location $\bfp_i\in\Re^3$ where it should be carried out is defined. Based on this, we introduce the binary variable $D_{i,k}$ which is equal to $1$ if the execution locations of task $\tau_i$ and $\tau_k$ are too close to be executed simultaneously, i.e., $D_{i,k} = 1$ if $\|\bfp_i-\bfp_k\|<\epsilon$ with $\epsilon$ a positive threshold, $0$ otherwise. These variables are symmetric by construction, i.e.,  $D_{i,k} = v $ implies $ D_{k,i}=v$ with $v\in\{0,1\}$, and enable to guarantee that no  tasks are simultaneously  carried out in the same location. Note that any criterion to define volume   occupation for executing each task  can be adopted to define $D_{i,k}$.  
\end{itemize}
Based on the estimated execution times, it is possible to define the maximum time $T_M$ to execute all the tasks. 
We assume that tasks are partitioned into groups or clusters according to common features, e.g., two pick-and-place operations of the same kind of object reasonably belong to the same cluster. We denote  the group that is associated with task~$\tau_i$ by $g(\tau_i)$.  
Note that this does not undermine the generality of the approach as $m$ different groups can be defined, i.e., each task can be associated with a different group.
In addition, we consider the following features for the tasks: 
\begin{enumerate}
\renewcommand{\labelenumi}{\text{\theenumi}}
\renewcommand{\theenumi}{F.\arabic{enumi}}
	\item \label{f:supervision} a task must be executed with a certain level of accuracy and may be supervised by a human operator to guarantee its correctness;
	\item \label{f:collaborative} a task may be required to  be carried out in a collaborative fashion, e.g., transporting heavy/large objects;
	\item \label{f:onlyhuman} a task can be suitable for humans only, e.g.,  manipulating objects with difficult geometry or which are highly deformable;
	\item \label{f:onlyrobot} a task can be suitable for robots only, e.g., carrying very heavy objects or manipulating objects that can be dangerous for human operators.
\end{enumerate}
 To formalize~\ref{f:supervision}, we introduce an \textit{execution quality index} $q_{i,j}\in[0,1]$, $\forall \tau_i\in\mathcal{T},\,a_j\in\mathcal{A}$  which, for each pair  (agent~$a_j$, task~$\tau_i$), assesses the quality of the execution of task $\tau_i$ by  agent $a_j$. We assume that tasks in the same group are associated with  same execution quality index for each agent~$a_j$, i.e., $q_{i,j}=q_{k,j}$ if $g(\tau_i)=g(\tau_k)$, $\forall \tau_i,\tau_k\in\mathcal{T}$. 
A minimum quality $\underline{q}$ is required to guarantee a minimum task execution accuracy. 
If an agent does not meet the minimum quality requirement for a certain task, a human operator can be assigned for its supervision. In this way, the human can monitor if the task execution is correct or not, and possibly intervene if necessary. We thus introduce a \emph{supervision quality} $q_{i,j}^s\in[0,1]$, $\forall \tau_i\in\mathcal{T},\,a_j\in\mathcal{A}$  which quantifies for each  human agent $a_j$ the achievable quality for task $\tau_i$ 
under his/her supervision. Human intervention is allowed during supervision, and supervision quality is equal to $0$ for robotic agents, i.e., only human operators can play supervision role.  
As for the execution quality index, we consider that 
tasks in the same group are associated with  same supervision quality index for each human agent~$a_{h,j}$, i.e., $q_{i,j}^s=q_{k,j}^s$ if $g(\tau_i)=g(\tau_k)$, $\forall \tau_i,\tau_k\in\mathcal{T}$. 
We consider that quality indices are additive, i.e.,  the overall quality of a task is given by the sum of the qualities of the agents that execute  and supervise it. Note that this is only a possible choice made in the paper, but the proposed framework could be easily extended to tackle different models for the quality, e.g., considering minimum or maximum quality of the involved agents as overall quality of the task.    
Note that the quality indices can generally vary over time and we update them at the end of the execution of each task as detailed in Section~\ref{sec:reall}. 
 
Concerning the remaining features, collaborative tasks in \ref{f:collaborative} are denoted by the binary variable $C_i,\,\forall \tau_i\in\mathcal{T}$ which is equal to $1$ if task~$\tau_i$ requires two agents to be accomplished, $0$ otherwise. With regard to  features~\ref{f:onlyhuman} and \ref{f:onlyrobot}, let $\mathcal{T}_j$ be the set of tasks that can be carried out by agent $a_j$ ( either robotic or human); these features can be easily expressed by setting $\Delta_{i,j}=M$, with $M$ an arbitrary high constant,  or $w_{i,j}=M$ for tasks $\tau_i \in \mathcal{T}\setminus \mathcal{T}_j$.

\section{Human multi-robot task allocation problem}
Let $X_{i,j}$ be the binary decision variable for the assignment of task $\tau_i$ to agent $a_j$, i.e.,  $X_{i,j}=1$ if  agent $a_j$ has to execute  task $\tau_i$, $X_{i,j}=0$ otherwise, let $S_{i,j}$ be the binary decision variable for the supervision of task $\tau_i$ by the human agent $a_j$, i.e., $S_{i,j}=1$ if  human $a_j$  has to supervise the execution of  task $\tau_i$, $S_{i,j}=0$ otherwise, and let $\underline{t}_{i},\overline{t}_{i}$ be the starting and end time of each task $\tau_i\in\mathcal{T}$. 
Our objective is to define the tasks assigned to each agent and the respective starting and end times as well as the supervision assignments by minimizing a cost function while complying with system constraints. In detail, the following MILP problem is formulated: 

\begin{subequations}  \label{eq:prob}
 \begin{align} 
 \label{eq:obj}
&\min_{X_{i,j}, S_{i,j},\underline{t}_i, \overline{t}_i} \underbrace{\max_{\tau_i\in\mathcal{T}} \frac{\overline{t}_i}{T_M}}_{\text{makespan}}-\sum_{\tau_i\in \mathcal{T}}\sum_{a_j\in \mathcal{A}}
\begin{aligned}[t]
&\big(\underbrace{q_{i,j} X_{i,j}+q^s_{i,j} S_{i,j}  }_{\text{overall quality}}\\
&-\underbrace{w_{i,j}X_{i,j}-w^s_{i,j}S_{i,j} }_{\text{overall workload}}\big)  
\end{aligned}\span\span\span\\
& \quad\quad \text{s.t.}\nonumber \\
&\sum_{a_j\in A}X_{i,j}=1+C_i, && \forall \tau_i\in\mathcal{T} &  \label{eq:task-ass}\\
&S_{i,j}+ X_{i,j} \leq 1, && \forall \tau_i\in\mathcal{T},a_{h,j}\in \mathcal{A}_h  &  \label{eq:supervision}\\
& \sum_{\tau_i \in \mathcal{T}}\sum_{a_{r,j} \in \mathcal{A}_r} S_{i,j} = 0  \span\span\span  \label{eq:sij}\\
&\sum_{a_j\in \mathcal{A}}\left(q_{i,j} X_{i,j}+q^s_{i,j} S_{i,j} \right)\geq \underline{q}, && \forall \tau_i\in\mathcal{T} &   \label{eq:qual}\\
& \overline{t}_i -\underline{t}_i \geq  X_{i,j}{\Delta}_{i,j }  && \forall \tau_i\in \mathcal{T}, a_j\in\mathcal{A}& \label{eq:duration}\\  
&\underline{t}_k -P_{i,k}\overline{t}_i \geq 0  && \forall \tau_i,\tau_k\in \mathcal{T} & \label{eq:duration:1}\\  
&
\begin{aligned}
&{\underline{t}_k-\overline{t}_i} \ge -M(2-X_{i,j}-X_{k,j} \\
&\quad - S_{i,j} - S_{k,j} ) -M(1-V_{i,k,j})
\end{aligned}
 && \forall\tau_i,\tau_k\in\mathcal{T}, a_j\in\mathcal{A} &\label{eq:agent-all}\\
 &
 \begin{aligned}
 &{\underline{t}_i-\overline{t}_k} \ge -M(2-X_{i,j}-X_{k,j} \\
 &\quad- S_{i,j} - S_{k,j} ) -M\,V_{i,k,j}
\end{aligned}
 && \forall \tau_i,\tau_k\in\mathcal{T},  a_j\in\mathcal{A} &\label{eq:agent-all-2}\\
&
\begin{aligned}
&{\underline{t}_k-\overline{t}_i} \ge -M(1-D_{i,k}) \\
& \quad\quad\quad\quad -M(1-Z_{i,k}) 
\end{aligned}
&& \forall\tau_i,\tau_k\in\mathcal{T} &\label{eq:spatial-all}\\
&
\begin{aligned}
&{\underline{t}_i-\overline{t}_k} \ge -M(1-D_{i,k}) \\
&\quad\quad\quad\quad-M\,Z_{i,k}
\end{aligned}
&& \forall\tau_i,\tau_k\in\mathcal{T} &\label{eq:spatial-all-2}
\end{align}
\end{subequations}
According to the objective function in~\eqref{eq:obj}, we  aim to minimize the system makespan, i.e., the overall execution time, normalized with respect to the maximum execution time $T_M$ while maximizing the overall process quality, given by the cumulative quality of each task, and minimizing the overall agents workload. The quality values at the allocation time are considered, and  supervision quality and workload by human operators are also taken into account. 
The following constraints are considered: 
\begin{itemize}
    \item equation~\eqref{eq:task-ass} ensures that the exact number of agents  is allocated to each task, i.e., for each task $\tau_i\in\mathcal{T}$ a total of one or two agents must execute it depending on whether the task is collaborative (one agent case) or not  (two agents case);
    \item equation~\eqref{eq:supervision} imposes the mutual exclusivity for supervision and execution by human operators, i.e., for each task $\tau_i\in\mathcal{T}$ a human agent cannot simultaneously execute ($X_{i,j}=1$) and supervise it ($S_{i,j}=1$); 
    \item  equation \eqref{eq:sij} imposes that no supervision can be performed by robotic agents $a_{r,j}\in\mathcal{A}_r$;
    \item equation \eqref{eq:qual} implies that a minimum quality $\underline{q}$ is guaranteed for each task $\tau_i\in\mathcal{T}$. The overall quality takes into account both the execution quality indices $q_{i,j}$ of the assigned agents with $X_{i,j}=1$ and the supervision quality indices $q^s_{i,j}$ for the assigned human supervisors with $S_{i,j}=1$; 
    \item  equation \eqref{eq:duration} defines the minimum duration of each task, i.e., if task $\tau_i$ is assigned to agent $a_j$ ($X_{i,j}=1$), then the allocation time $\overline{t}_i-\underline{t}_i$ must be at least equal to $\Delta_{i,j}$;    
    \item equation \eqref{eq:duration:1} allows to guarantee the required tasks sequentiality, i.e., for each pair of tasks $\tau_i,\tau_k\in\mathcal{T}$ if $P_{i,k}=1$ then $\tau_k$ has to start after $\tau_i$ ends, while if $P_{i,k}=0$ no precedence is imposed;  
    \item equations \eqref{eq:agent-all}-\eqref{eq:agent-all-2} ensure that each agent $a_j\in\mathcal{A}$ cannot
    simultaneously execute or supervise more than one task. To this aim, the auxiliary decision binary variable $V_{i,k,j}$ $\forall a_j\in\mathcal{A},\tau_i,\tau_k\in\mathcal{T}$ is introduced. Let us consider the case in which both tasks $\tau_i,\tau_k$ are assigned to agent $a_j$, i.e., $X_{i,j}=X_{k,j}=1$. By virtue of \eqref{eq:supervision} and \eqref{eq:sij}, it follows that $S_{i,j}=S_{k,j}=0$. Equations \eqref{eq:agent-all}-\eqref{eq:agent-all-2} thus lead to 
    $$
    \begin{aligned}
    \underline{t}_k-\overline{t}_i &\ge  -M(1-V_{i,k,j})\\
    {\underline{t}_i-\overline{t}_k} &\ge -M\,V_{i,k,j}
    \end{aligned}
    $$
    implying that either $\underline{t}_k\geq\overline{t}_i$ (if $V_{i,k,j}=1$) or $\underline{t}_i\geq\overline{t}_k$ (if $V_{i,k,j}=0$), but no simultaneous execution can occur. Similar reasoning applies for the case in which  $S_{i,j}=S_{k,j}=1$. Finally, when the tasks are not assigned to the same robots, no constraints are imposed by \eqref{eq:agent-all}-\eqref{eq:agent-all-2}; 
    \item equations \eqref{eq:spatial-all}-\eqref{eq:spatial-all-2} allow to specify that tasks that occupy the same spatial location are not executed simultaneously. Similarly to the constraints in  \eqref{eq:agent-all}-\eqref{eq:agent-all-2}, we introduce an auxiliary  decision binary variable $Z_{i,k}$ $\forall \tau_i,\tau_k\in\mathcal{T}$. Based on this, if a spatial limitation exists between tasks $\tau_i,\tau_k$, then equations \eqref{eq:spatial-all}-\eqref{eq:spatial-all-2} lead to specify that either $\tau_k$ starts after $\tau_i$, i.e.,  $\underline{t}_k\geq\overline{t}_i$ when $Z_{i,k}=1$, or the opposite holds true, i.e.,  $\underline{t}_i\geq\overline{t}_k$ when $Z_{i,k}=0$. 
\end{itemize}
 Note that the proposed formulation is particularly versatile and can be easily adapted to different collaborative production processes and tasks involving an arbitrary number of humans and robots. 

\section{Online Re-allocation}\label{sec:reall}

Due to uncertainty in realistic scenarios, 
online re-allocation may be necessary to meet future plan constraints, as well as to preserve the optimality of the plans to a certain extent. For this reason, we consider that at the end of the execution of each task, first a \emph{monitoring} step is performed to update the respective parameters, i.e.,  quality, execution time, and workload, next, the evaluation of a \emph{re-allocation condition}  establishes whether re-allocation is necessary or not for future plans on the basis of the updated parameters.  
Note that re-allocation should only be performed when necessary to avoid excessive computational burden of solving problem \eqref{eq:prob} and the mental overload  caused by frequent task switching~\cite{Wickens_IJHCS2015}. 

\subsubsection{Parameters monitoring and update}

The first step for re-allocation is to measure the quality and workload indices and execution times of both human  and robotic agents at the end of the execution of each task~$\tau_i$. We refer to the values of the parameters before the update as \emph{nominal} values in the following. Based on these measures, the parameters are updated as follows. 

Concerning the quality update, 
a distinction is made depending on whether the task~$\tau_i$ is supervised or not. 
In the case no supervision is foreseen, i.e., $S_{i,j}=0$ $\forall a_{h,j}\in\mathcal{A}_h$,  the measured quality becomes the current execution quality for the assigned agents. When the task is collaborative, i.e., $C_i=1$, the measured quality is equally distributed between the two agents. In addition, when the quality index $q_{i,j}$ is updated, also the quality indices of agent~$a_j$ for tasks belonging to the same group~$g(\tau_i)$ are updated accordingly, i.e., $q_{k,j}=q_{i,j}$ $\forall \tau_k \in\mathcal{T}$ such that $g(\tau_i)=g(\tau_k)$. In the case supervision is foreseen, i.e., it exists $a_{h,j}\in\mathcal{A}_h$ such that $S_{i,j}=1$,
 we further distinguish the cases in which the human does and does not intervene during the supervision. This is motivated by the fact that when human supervision is required, the human does not need to necessarily intervene during the task execution if the assigned agents are able to carry it out autonomously. Hence, if the human does not intervene, the measured quality becomes the execution quality of the executing agents, while if the human intervenes,  the measured quality  becomes his/her supervision quality and no update is made on the execution quality of the assigned agents. 
 
As far as the workload and the execution times are concerned, they are updated to the measured values. Moreover, workload and the execution times for the tasks of the same group~$g(\tau_i)$ are updated according to the same proportion, e.g., if 
the measured execution time is increased by $10\%$ compared to the nominal one, then also the execution times of the tasks belonging to the same group are increased by $10\%$. Clearly, only the parameters associated with the involved agents, either supervising or executing the task, are updated. Furthermore, any other update policy is possible depending on the particular scenario.

\subsubsection{Re-allocation strategy}
After the parameters are updated, it is necessary to establish whether to online re-allocate the remaining tasks or not. We propose to perform re-allocation  \emph{i)} firstly, to ensure the feasibility of the allocation, i.e., re-allocation is performed if constraints of future tasks are violated considering the updated parameters. As instance,  if the execution quality of an agent for a group $g(\tau_i)$ is decreased to a value lower than $\underline{q}$ during the  monitoring and update step, and a task from this group must be executed by the same agent in the future, then supervision may be required to guarantee the minimum quality constraint~\eqref{eq:qual}; \emph{ii)} secondly, 
re-allocation depends on  the change in performance. More specifically, let  $\mathcal{T}^+\subset \mathcal{T}$ be the  subset of tasks which still need to be executed and let $\hat{C}^+$ be the cost function related to tasks  in $\mathcal{T}^+$ and evaluated with the parameters available at current planning time, i.e., 
$$
\begin{aligned}
\hat{C}^+ = \displaystyle\max_{\tau_i\in\mathcal{T}^+} \frac{\overline{t}_i}{T_M}-\displaystyle\sum_{\tau_i\in \mathcal{T}^+}\displaystyle\sum_{a_j\in \mathcal{A}}&\Big({\hat{q}_{i,j} X_{i,j}+\hat{q}^s_{i,j} S_{i,j}  }\\ &-{\hat{w}_{i,j}X_{i,j}-\hat{w}^s_{i,j}S_{i,j} }\Big)  
\end{aligned}
$$
where the notation $\hat{(\cdot)}$ is used to denote the best estimate of the parameters available so far. Similarly, let ${C}^+$ be the cost function related to future tasks and evaluated with the \emph{updated} parameters. Re-allocation is carried out when the following condition is met
\begin{equation}\label{eq:delta}
    \delta \triangleq \frac{|\hat{C}^+ - {C}^+|}{\hat{C}^+}>\delta_t
\end{equation}
with $\delta_t$ a positive constant. The rationale behind~\eqref{eq:delta} is that re-allocation is also performed to preserve the allocation optimality with a certain tolerance. In particular, the higher the re-allocation parameter~$\delta$, the more likely the allocation made with parameters at planning time is not optimal for the updated parameters. 
Finally, online re-allocation is performed also \emph{iii)} when a new batch of operations to execute is available or \emph{iv)} when the available resources, either robotic or human, change with respect to the planned ones~\cite{NIKOLAKIS2018237}, e.g., a new human operator is available or a fault on a robot occurs.

\section{Simulation results}
In this section, the proposed approach is evaluated in a realistic simulation environment.

\subsection{Simulation setup}
\definecolor{darkgreen}{rgb}{0, 0.38, 0.02}
\begin{psfrags}
  \def\scal{0.9}  
  \def\scalsmall{0.7}  
  \psfrag{des}[cc][][\scalsmall]{\textcolor{darkgreen}{Desired structure}}
  \psfrag{r1}[cc][][\scal]{$a_{r,1}$}
  \psfrag{h1}[cc][][\scal]{$a_{h,1}$}
  \psfrag{r2}[cc][][\scal]{$a_{r,2}$}
  \psfrag{1}[cc][][\scalsmall]{$1$}
  \psfrag{2}[cc][][\scalsmall]{$2$}
  \psfrag{3}[cc][][\scalsmall]{$3$}
  \psfrag{4}[cc][][\scalsmall]{$4$}
  \psfrag{5}[cc][][\scalsmall]{$5$}
  \psfrag{6}[cc][][\scalsmall]{$6$}
  \psfrag{7}[cc][][\scalsmall]{$7$}
  \psfrag{8}[cc][][\scalsmall]{$8$}
  \psfrag{9}[cc][][\scalsmall]{$9$}
  \psfrag{10}[cc][][\scalsmall]{${10}$}
  \psfrag{11}[cc][][\scalsmall]{${11}$}
  \psfrag{12}[cc][][\scalsmall]{${12}$}
  \psfrag{13}[cc][][\scalsmall]{${13}$}
  \psfrag{14}[cc][][\scalsmall]{${14}$}
  \psfrag{x}[cc][][\scalsmall]{$x$}
  \psfrag{y}[cc][][\scalsmall]{$y$}
  \psfrag{z}[cc][][\scalsmall]{$z$}
\mypsfrag{8.5}{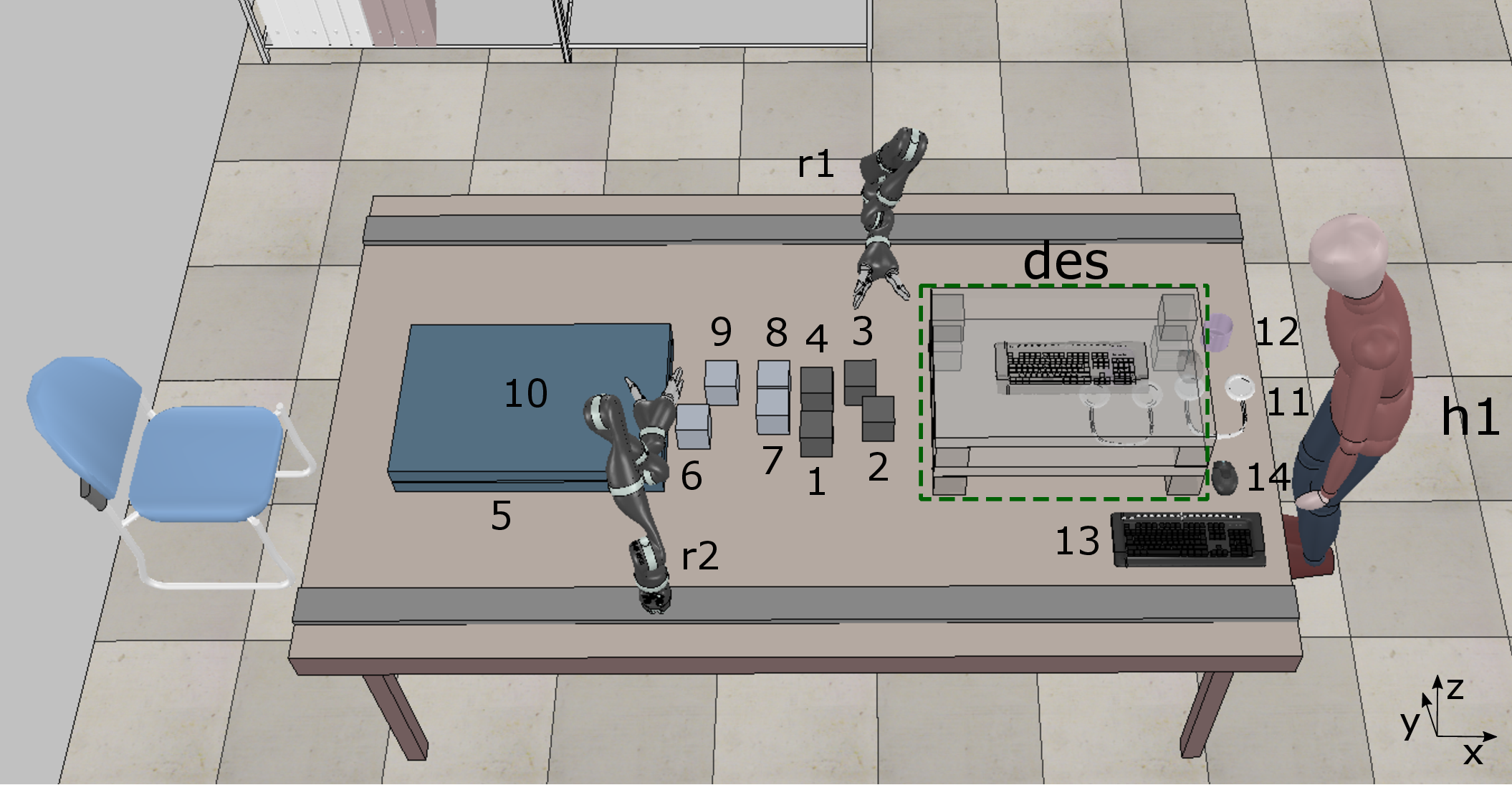}{-10pt}{Simulation setup composed of two manipulators ($a_{r,1}$,$a_{r,2}$) and a human operator ($a_{h,1}$). Objects to assemble are numbered and the desired structure is shown in transparency in the dashed box. The reference frame is shown in the bottom right corner.    }{fig:setup}
\end{psfrags}

\begin{psfrags}
  \def\scal{0.8}  
  \def\scalsmall{0.7}  
  \psfrag{mat}[cc][][\scal]{\shortstack[c]{Matlab with \\ Gurobi solver}}
  \psfrag{vrep}[cc][][\scal]{\shortstack[c]{V-REP \\ simulation}}
  \psfrag{c1}[cc][][\scal]{visual feedback}
  \psfrag{c2}[cc][][\scal]{commands}
  \psfrag{c3}[cc][][\scal]{sends}
  \psfrag{c4}[cc][][\scal]{commands}
  \psfrag{gr}[cc][][\scalsmall]{\shortstack[c]{Grasp/ \\ release}}
  \psfrag{z}[cc][][\scalsmall]{\shortstack[c]{Motion \\ on $z$}}
  \psfrag{xy}[cc][][\scalsmall]{\shortstack[c]{Motion \\ on $x$ and $y$}}
\mypsfrag{8.5}{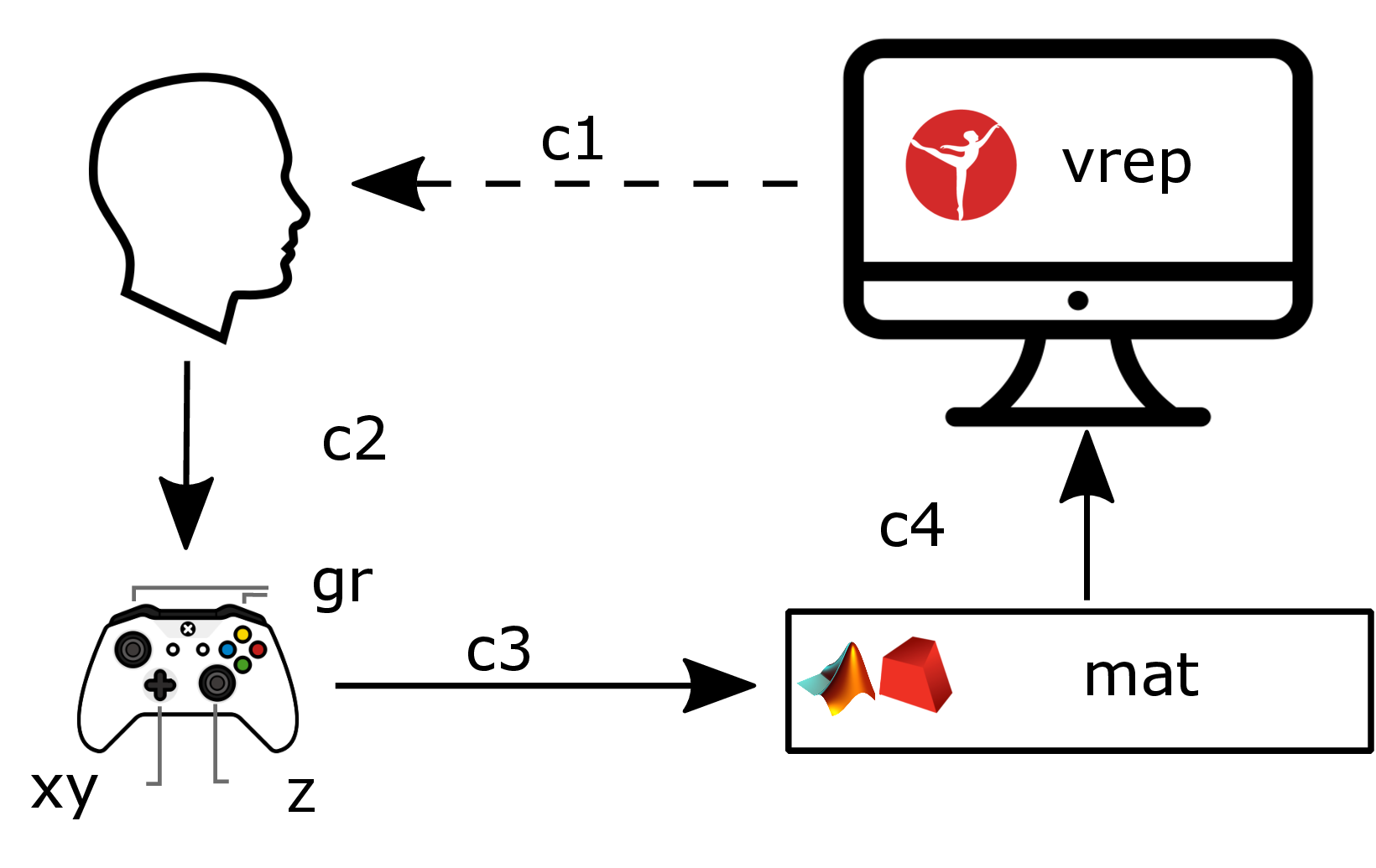}{-10pt}{Architecture of the validation setup. The human operator sends commands through a controller which is interfaced with Matlab. The latter resorts to  Gurobi solver to find a solution for the optimization problem and communicates with V-REP simulation environment. Visual feedback is provided to the human through V-REP.  }{fig:architecture}
\end{psfrags}

A simulation collaborative setup, shown in \Cref{fig:setup}, composed of two Kinova Jaco$2$ ($n_r=2$), with $7$-DOFs, mounted on sliding tracks ($1$-DOF) and a human operator ($n_h=1$) is employed to validate  the proposed approach.  The left human hand of the simulated human operator is teleoperated in real-time by a person through a Microsoft Xbox controller.  The developed architecture is reported in \Cref{fig:architecture}. 
Specific buttons are used to grasp/release objects as well as to move along $x$,$y$ and $z$ axes as detailed in \Cref{fig:architecture} (bottom left). 
All software components are developed in Matlab interfaced with \emph{i)} the controller, to receive human inputs, \emph{ii)} Gurobi solver\footnote{https://www.gurobi.com}, to solve the optimization problem, and \emph{iii)} Coppelia V-REP\footnote{https://www.coppeliarobotics.com}, to simulate the operating environment. The latter provides  visual feedback  to the human operator. An illustrative video showing the effectiveness of the developed framework is available at the link\footnote{\label{fn:video}\url{http://webuser.unicas.it/lai/robotica/video/HRC-MILP.mp4}}.

A collaborative assembly process is considered in which a structure with two levels is built, as depicted in transparency in \Cref{fig:setup} and highlighted by the green dashed box.
In particular, the following objects, divided in three groups, are involved: \emph{i)} $8$ cubes that make up the bases of the levels (four cubes for each level), \emph{ii)}  two planar surfaces with size \mbox{$0.75$ m $\times\, 0.5$ m $ \times\, 0.05$ m} (one for each level), \emph{iii)} four office items, namely a cup, headphones, keyboard and mouse, which are added on the planar surfaces (two on each level).  

Objects are numbered as shown in \Cref{fig:setup} and a pick-and-place task $\tau_i$ is defined for each object~$i$. Due to the size of the planar surfaces (objects $5$ and $10$), their transport cannot be carried out by a single agent. Therefore, the respective tasks are set as collaborative, i.e., $C_5=C_{10}=1$. 

Precedence constraints are introduced to properly build the two-level structure: tasks~$1-4$, which position the cubic bases for the bottom level, need to be completed in order to start task~$10$, which positions the bottom planar surface; the latter task then needs to be finished to position the cubic bases for the top level (tasks~$6-9$) as well as the office items~$11,12$. After tasks~$6-9$ and~$11,12$ are accomplished, the top planar surface can be placed (task~$5$). Finally, when the planar surface is mounted, the remaining two office items (tasks~$13-14$) can be placed. 

Spatial constraints are defined for the cubic bases~$7,8$, i.e., $D_{7,8}=D_{8,7}=1$, since they are placed close together in the initial configuration as shown in~\Cref{fig:setup}. For the same reason, spatial constraints are also introduced for the office items~$13,14$ (keyboard and mouse),  i.e., \mbox{$D_{13,14}=D_{14,13}=1$}.

Execution times for the robotic agents $a_{r,1}, a_{r,2}$ 
are computed by considering average linear velocity equal  to $0.05$~m/s and average angular velocity equal to $1.3$~rad/s. Execution times for the human agent are set by recording once the required times by the human operator to perform the tasks and multiplying these times  by a factor greater than $1$ ($1.1$ in our case) in order to obtain more conservative estimates. 
Moreover, $\Delta_{i,j}=M$ is set if object~$i$ is outside the reachable workspace of agent~$j$. The resulting maximum execution time in \eqref{eq:obj} is $T_M = 668.37$~s. 

Finally,  workload of the robotic agents ($a_1,a_2\in\mathcal{A}$) is initialized to $0.5$ for tasks associated with the first group of objects, i.e., for the cubic bases of the structure, while it is set equal to~$1$ for the remaining tasks. Concerning the human agent, unit workload is considered for executing  and supervising tasks associated with groups $1$ and $3$, i.e., \mbox{$w_{i,3}=1$}, $w^s_{i,3}=1$ for each $\forall \tau_i$ such that $g(\tau_i) = 1$ or $g(\tau_i) = 3$,  with  $3$ the index of the human agent (i.e., $a_3\in \mathcal{A}$). High execution workload is considered instead to perform the collaborative tasks~$5$ and~$10$ for which it holds $w_{5,3}=w_{10,3}=M$, thus preventing the assignment of these heavy tasks to the human operator. 

The following set of parameters is used: $\underline{q}=0.8$ and $M=1000$ in the problem formulation~\eqref{eq:prob} and $\delta_t=0.15$ for the re-allocation condition in~\eqref{eq:delta}. 

In the following, first we  discuss  the allocation and results obtained to carry out the described assembly process with the proposed framework. Next, we perform a simulation campaign to prove the effectiveness of the re-allocation strategy. Quality values are detailed in the respective case studies.

\subsection{Experimental results}

\begin{psfrags}
  \def\scal{0.8}  
  \def\scalnum{0.6}  
  \def\scalsmall{0.8}  
  \psfrag{t}[cc][][\scal]{$t$[s]}
  \psfrag{a}[cc][][\scal]{$a$}
  \psfrag{t1}[cc][][\scalsmall]{$\tau_1$}
  \psfrag{t2}[cc][][\scalsmall]{$\tau_2$}
  \psfrag{t3}[cc][][\scalsmall]{$\tau_3$}
  \psfrag{t4}[cc][][\scalsmall]{$\tau_4$}
  \psfrag{t5}[cc][][\scalsmall]{$\tau_5$}
  \psfrag{t6}[cc][][\scalsmall]{$\tau_6$}
  \psfrag{t7}[cc][][\scalsmall]{$\tau_7$}
  \psfrag{t8}[cc][][\scalsmall]{$\tau_8$}
  \psfrag{t9}[cc][][\scalsmall]{$\tau_9$}
  \psfrag{t10}[cc][][\scalsmall]{$\tau_{10}$}
  \psfrag{t11}[cc][][\scalsmall]{$\tau_{11}$}
  \psfrag{t12}[cc][][\scalsmall]{$\tau_{12}$}
  \psfrag{t13}[cc][][\scalsmall]{$\tau_{13}$}
  \psfrag{t14}[cc][][\scalsmall]{$\tau_{14}$}
  \psfrag{-100}[cc][][\scalnum]{ $-100$}
\psfrag{-50}[cc][][\scalnum]{ $-50$}
\psfrag{-40}[cc][][\scalnum]{ $-40$}
\psfrag{-30}[cc][][\scalnum]{ $-30$}
\psfrag{-20}[cc][][\scalnum]{ $-20$}
\psfrag{-4}[cc][][\scalnum]{ $-4$}
\psfrag{-2}[cc][][\scalnum]{ $-2$}
\psfrag{-0.4}[cc][][\scalnum]{ $-0.4$}
\psfrag{-0.3}[cc][][\scalnum]{ $-0.3$}
\psfrag{-0.2}[cc][][\scalnum]{ $-0.2$}
\psfrag{-0.1}[cc][][\scalnum]{ $-0.1$}
\psfrag{0}[cc][][\scalnum]{ $0$}
\psfrag{0.002}[cc][][\scalnum]{$0.002$}
\psfrag{0.004}[cc][][\scalnum]{$0.004$}
\psfrag{0.006}[cc][][\scalnum]{$0.006$}
\psfrag{0.008}[cc][][\scalnum]{$0.008$}
\psfrag{0.005}[cc][][\scalnum]{ $0.005$}
\psfrag{0.01}[cc][][\scalnum]{ $0.01$}
\psfrag{0.012}[cc][][\scalnum]{ $0.012$}
\psfrag{0.015}[cc][][\scalnum]{ $0.015$}
\psfrag{0.02}[cc][][\scalnum]{$0.02$}	
\psfrag{0.025}[cc][][\scalnum]{$0.025$}	
\psfrag{0.04}[cc][][\scalnum]{$0.04$}	
\psfrag{0.06}[cc][][\scalnum]{$0.06$}	
\psfrag{0.08}[cc][][\scalnum]{$0.08$}
\psfrag{0.05}[cc][][\scalnum]{ $0.05$}
\psfrag{0.1}[cc][][\scalnum]{ $0.1$}
\psfrag{0.2}[cc][][\scalnum]{ $0.2$}
\psfrag{0.4}[cc][][\scalnum]{ $0.4$}
\psfrag{0.6}[cc][][\scalnum]{ $0.6$}
\psfrag{0.8}[cc][][\scalnum]{ $0.8$}
\psfrag{0.82}[cc][][\scalnum]{ $0.82$}
\psfrag{0.83}[cc][][\scalnum]{ $0.83$}
\psfrag{0.84}[cc][][\scalnum]{ $0.84$}
\psfrag{0.85}[cc][][\scalnum]{ $0.85$}
\psfrag{0.86}[cc][][\scalnum]{ $0.86$}
\psfrag{0.5}[cc][][\scalnum]{ $0.5$}	\psfrag{1}[cc][][\scalnum]{ $1$}
\psfrag{1.5}[cc][][\scalnum]{ $1.5$}
\psfrag{2}[cc][][\scalnum]{ $2$}
\psfrag{2.5}[cc][][\scalnum]{ $2.5$}
\psfrag{3}[cc][][\scalnum]{ $3$}
\psfrag{4}[cc][][\scalnum]{ $4$}
\psfrag{4.5}[cc][][\scalnum]{ $4.5$}
\psfrag{5}[cc][][\scalnum]{ $5$}
\psfrag{5.2}[cc][][\scalnum]{ $5.2$}
\psfrag{5.4}[cc][][\scalnum]{ $5.4$}
\psfrag{5.6}[cc][][\scalnum]{ $5.6$}
\psfrag{5.8}[cc][][\scalnum]{ $5.8$}
\psfrag{6}[cc][][\scalnum]{ $6$}
\psfrag{7}[cc][][\scalnum]{ $7$}
\psfrag{8}[cc][][\scalnum]{ $8$}
\psfrag{9}[cc][][\scalnum]{ $9$}
\psfrag{10}[cc][][\scalnum]{ $10$}
\psfrag{11}[cc][][\scalnum]{ $11$}
\psfrag{12}[cc][][\scalnum]{ $12$}
\psfrag{13}[cc][][\scalnum]{ $13$}
\psfrag{14}[cc][][\scalnum]{ $14$}
\psfrag{15}[cc][][\scalnum]{ $15$}
\psfrag{16}[cc][][\scalnum]{ $16$}
\psfrag{18}[cc][][\scalnum]{ $18$}
\psfrag{20}[cc][][\scalnum]{ $20$}
\psfrag{25}[cc][][\scalnum]{ $25$}
\psfrag{26}[cc][][\scalnum]{ $26$}
\psfrag{28}[cc][][\scalnum]{ $28$}
\psfrag{30}[cc][][\scalnum]{ $30$}
\psfrag{32}[cc][][\scalnum]{ $32$}
\psfrag{34}[cc][][\scalnum]{ $34$}
\psfrag{35}[cc][][\scalnum]{ $35$}
\psfrag{36}[cc][][\scalnum]{ $36$}
\psfrag{38}[cc][][\scalnum]{ $38$}
\psfrag{40}[cc][][\scalnum]{ $40$}
\psfrag{50}[cc][][\scalnum]{ $50$}
\psfrag{60}[cc][][\scalnum]{$60$}
\psfrag{70}[cc][][\scalnum]{$70$}
\psfrag{80}[cc][][\scalnum]{$80$}
\psfrag{90}[cc][][\scalnum]{$90$}
\psfrag{100}[cc][][\scalnum]{$100$}
\psfrag{120}[cc][][\scalnum]{$120$}
\psfrag{140}[cc][][\scalnum]{$140$}
\psfrag{160}[cc][][\scalnum]{$160$}
\psfrag{150}[cc][][\scalnum]{$150$}
\psfrag{180}[cc][][\scalnum]{$180$}
\psfrag{200}[cc][][\scalnum]{$200$}
\psfrag{250}[cc][][\scalnum]{$250$}
\psfrag{300}[cc][][\scalnum]{$300$}
\psfrag{350}[cc][][\scalnum]{$350$}
\psfrag{400}[cc][][\scalnum]{$400$}
\psfrag{500}[cc][][\scalnum]{$500$}
\psfrag{600}[cc][][\scalnum]{$600$}
\psfrag{700}[cc][][\scalnum]{$700$}
\psfrag{800}[cc][][\scalnum]{$800$}
\psfrag{900}[cc][][\scalnum]{$900$}
\psfrag{1000}[cc][][\scalnum]{$1000$}
\psfrag{1200}[cc][][\scalnum]{$1200$}
\psfrag{1400}[cc][][\scalnum]{$1400$}
\psfrag{1600}[cc][][\scalnum]{$1600$}
\psfrag{1800}[cc][][\scalnum]{$1800$}
\psfrag{2000}[cc][][\scalnum]{$2000$}
\psfrag{-0.5}[cc][][\scalnum]{ $\!\!\!-0.5$}
\psfrag{-1}[cc][][\scalnum]{ $\!\!\!-1$}
\psfrag{-1.5}[cc][][\scalnum]{ $\!\!\!-1.5$}
\psfrag{-2.6}[cc][][\scalnum]{ $\!\!\!-2.6$}
\psfrag{-2.8}[cc][][\scalnum]{ $\!\!\!-2.8$}
\psfrag{-3}[cc][][\scalnum]{ $\!\!\!-3$}
\psfrag{-3.2}[cc][][\scalnum]{ $\!\!\!-3.2$}
\psfrag{-5}[cc][][\scalnum]{ $\!\!\!-5$}
\psfrag{-10}[cc][][\scalnum]{ $\!\!\!-10$}

\mypsfrag{8.5}{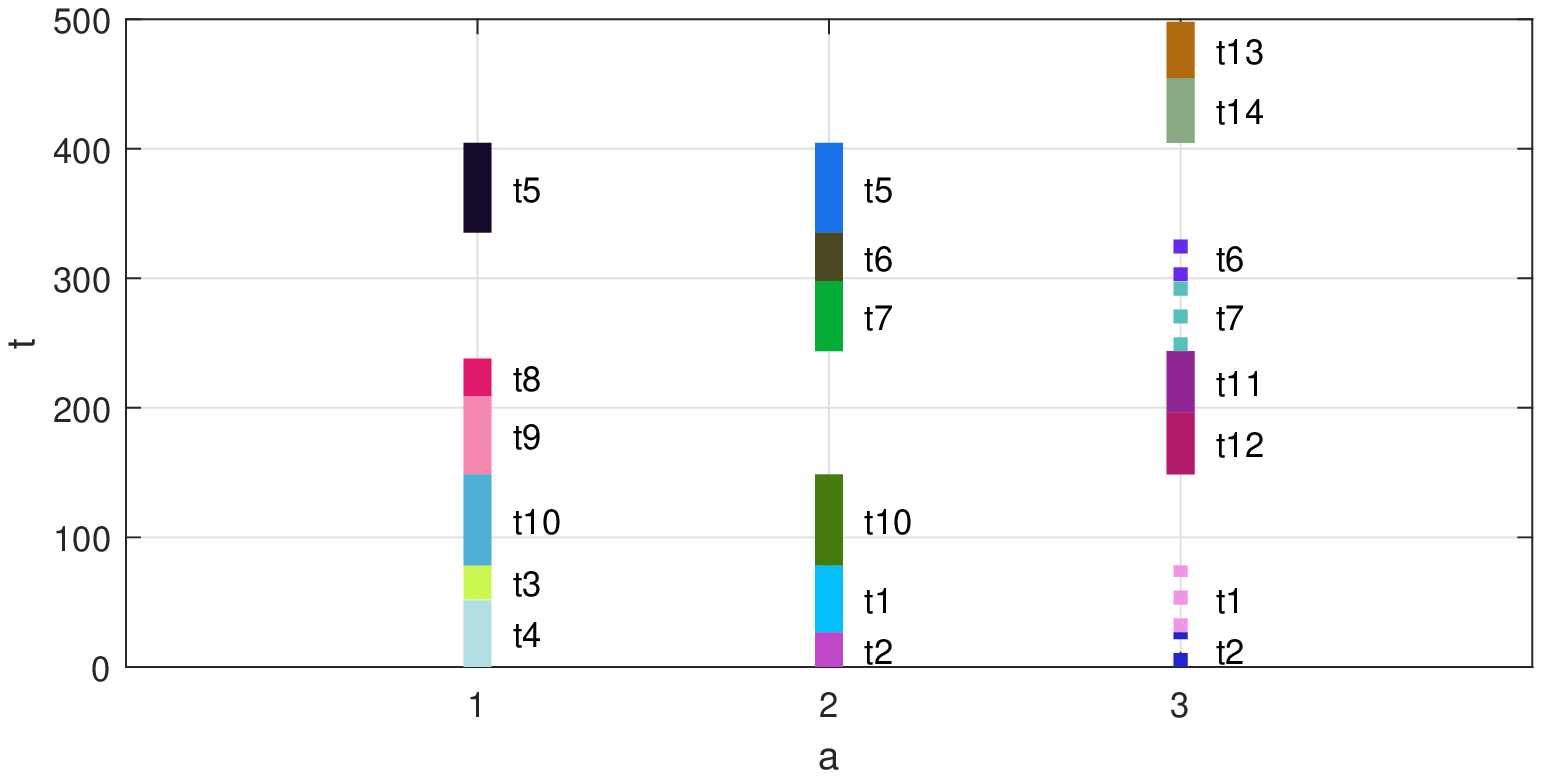}{-10pt}{ Optimal allocation for the three agents to perform the assembly process. Each task is highlighted with a different color and start and end times are represented as well as assigned agents. Dashed thin lines are used to denote supervision tasks by the human operator $a_3$.    }{fig:allocation}
\end{psfrags}

The experiment consists in performing the assembly process according to the optimal allocation obtained as solution of~\eqref{eq:prob}. 
  Quality indices for the two robotic agents are initialized to $0.8$ and $0.7$, respectively,  for tasks belonging to group~$1$ (i.e., cubic bases), to $0.4$ and $0.4$ for the collaborative tasks belonging to group~$2$ (i.e., planar surfaces), and to $0.5$ and $0.5$ for tasks belonging to group~$3$ (i.e., office items). The latter low values are motivated by the fact that the office items have more complex shapes to be manipulated by the robots than the objects in the other groups. 
  With regard to  the human agent, unit execution and supervision quality indices are considered for all tasks, i.e., \mbox{$q_{i,3}=1$}, $q^s_{i,3}=1$ $\forall \tau_i\in\mathcal{T}$.  

Before starting the execution of the tasks or when re-allocation is performed, the human operator is informed of the planned allocation through a visual information as shown, for example,  in \Cref{fig:allocation}. At execution time, whenever the person has to start executing or supervising a task, an appropriate message is displayed  in the V-REP simulator console as reported in the attached video together with a complete execution of the experiment\footref{fn:video}. In addition, an audio signal is generated to inform the human operator. 
When a task~$\tau_i$ is required to be supervised, i.e., $S_{i,3}=1$, the human operator can intervene and possibly reposition the object  involved in the task. 
Let $\mathcal{N}(\mu,\sigma^2)$ denote a Gaussian distribution with mean $\mu$ and variance~$\sigma^2$. To simulate realistic scenarios, a   perturbation of the final objects position generated according to a Gaussian distribution $\mathcal{N}(0,0.02)$ is introduced during the pick-and-place operations by the robotic agents.   
During the monitoring phase, the quality of a completed task is assessed by computing the difference between  the desired configuration of the object and its measured one. 

Figure~\ref{fig:allocation} shows the planned optimal allocation to carry out the assembly process. In detail, 
tasks (highlighted with different colors) assigned to each agent~$a_j$ with $j=1,2,3$ are represented along with their start and final times. Dashed thin lines are used to denote supervision tasks by the human operator $a_3$. 
A solution with makespan equal to $497.97$~s is found in which all precedence, quality, simultaneity  and spatial constraints are fulfilled. In particular, the execution of all tasks belonging to group~$1$ is distributed between the two robotic agents: tasks $3,4,8,9$ are assigned to agent~$a_1$,  while tasks $1,2,6,7$ are assigned to agent~$a_2$. Among the latter tasks,  task~$8$ starts after task~$9$ is accomplished in order to meet the spatial constraints. Moreover, supervision of the tasks $1,2,6,7$, executed by the robotic agent~$a_2$,  is obtained in order to ensure minimum quality~$\underline{q}$. In this regard, as visible from the video, the human operator only intervenes during task~$\tau_7$ to reposition the respective cubic base. Collaborative tasks~$5$ and $10$ are assigned to the robotic agents to position the planar surfaces, while office items (i.e., tasks $11,12,13,14$) are assigned to the human operator, achieving the best compromise quality/effort among the involved agents.  
Online monitoring of the parameters is made during the execution but no  constraints are violated nor the performance parameter $\delta$ exceeds the re-allocation threshold $\delta_t$, implying that no re-allocation is made in this case study. However, its validation is extensively carried out in the following.

\subsection{Re-allocation results}
To prove the effectiveness of the online re-allocation procedure, we perform a simulation campaign in which we monitor the values of the performance parameter $\delta$, quantifying the similarity between the planned cost and the measured one,  as well as of the cost function obtained with and without re-allocation. In particular, this case study focuses on evaluating the optimality provided by the re-allocation strategy compared to the static allocation. Random initial conditions as well as random online perturbations of the performance are considered for the simulation campaign. 
Initial values of the quality indices (both for execution and supervision) are set by adding a Gaussian random noise $\mathcal{N}(0,0.2)$ 
to the minimum quality~$\underline{q}$. The actual execution time of each task is generated by perturbing the value at planning time with a Gaussian random noise $\mathcal{N}(0,0.02)$. Similarly, at the end of the execution of each task, measured quality and workload for execution and supervision are generated by perturbing the values at planning time with a Gaussian random noise $\mathcal{N}(0,0.1)$. 
Perturbations are generated in such a way to always guarantee feasibility of the allocation, thus enabling the evaluation of the re-allocation impact  on the solution optimality. 
Moreover, possible human intervention during supervised tasks is established according to a uniform discrete distribution in the range~$\{0,1\}$.

\begin{psfrags}
  \def\scal{0.8}  
  \def\scalnum{0.6}  
  \def\scalsmall{0.6}  
  \psfrag{y}[cc][][\scal]{$\delta$}
  \psfrag{x}[cc][][\scal]{$\tau$}
  \psfrag{reall}[cc][][\scalsmall]{with re-all.}
  \psfrag{noreall}[cc][][\scalsmall]{without re-all.}
  
\mypsfrag{8.5}{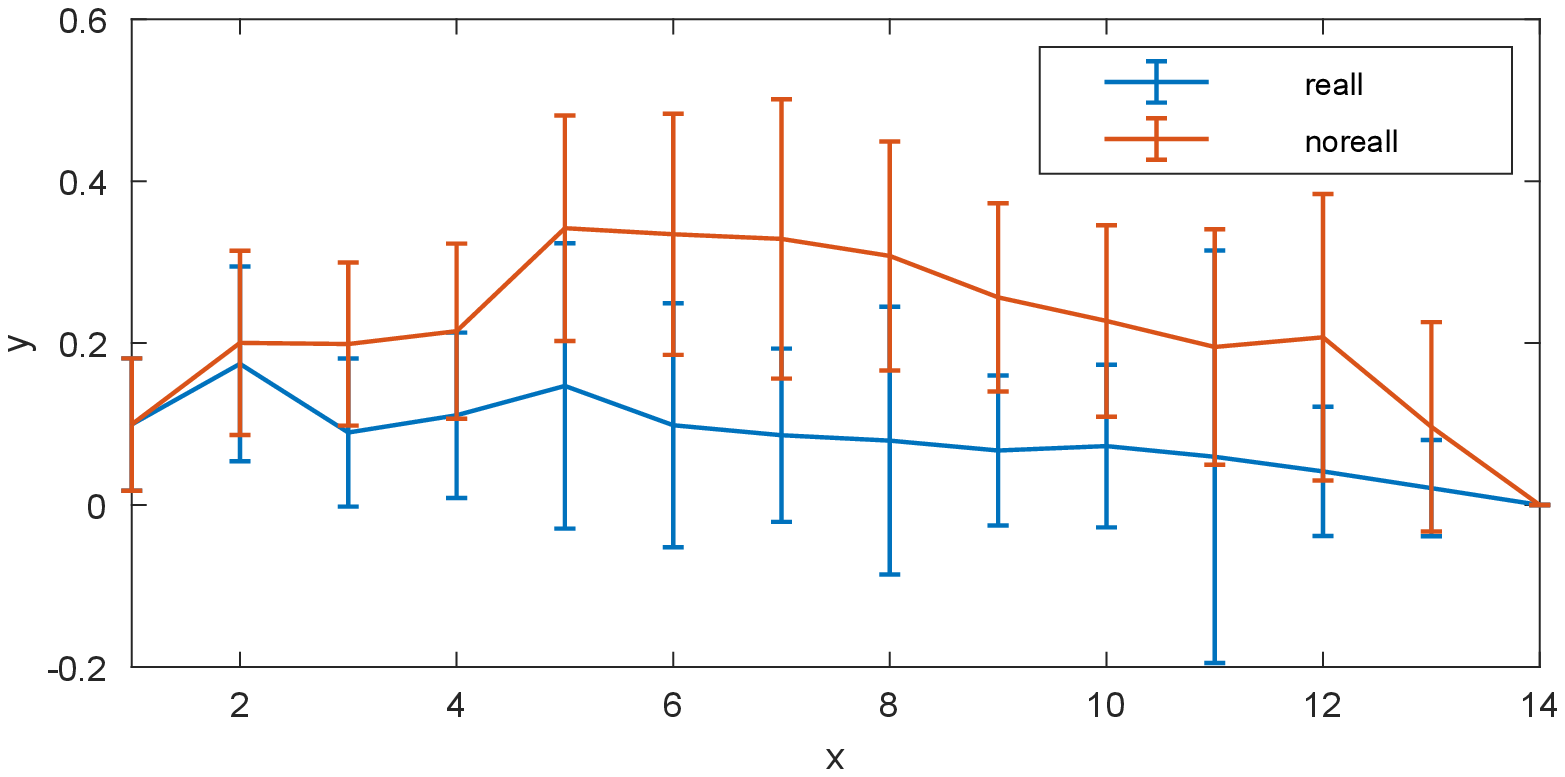}{-10pt}{ Comparison of the values of the parameter~$\delta$ at the end of each task obtained when re-allocation is enabled (in blue) and when it is not (in red). Average and standard deviation values are computed over $50$ trials.   }{fig:reallocation_plot}
\end{psfrags}

Figure~\ref{fig:reallocation_plot} reports the obtained values of the parameter~$\delta$ at the end of each task when re-allocation is enabled (in blue) and when it is not (in red), i.e., in the latter case the initial allocation is maintained throughout the execution. Average and standard deviation values over $50$ trials are shown. The figure makes evident the general improvement (i.e., lower values) given by the re-allocation procedure on the parameter~$\delta$ compared to the case of static allocation. More specifically,  average $\delta$ equal to  $0.08$ is achieved with reallocation, while average $\delta$ equal to $0.21$ is reached  without reallocation. Note that same results for the initial and final tasks are obtained since at the beginning the same initial allocation is considered by the two solutions (with and without re-allocation), while at the end no further tasks need to be executed meaning that $\mathcal{T}^+=\emptyset$ and $\hat{C}^+=C^+=0$.

Finally, the optimality improvement is also confirmed by the evaluation of the cost function achieved by the solutions with and without reallocation. In particular, the former achieves  $0.91 \pm 1.15$,  while the latter achieves \mbox{$2.62 \pm 0.68$}, thus leading to  an average improvement on the obtained cost function equal to  $\approx 65\%$ compared to the case of static allocation (i.e., without re-planning). 

\section{Conclusion}
In this work,  the task allocation problem in a human multi-robot collaborative scenario was addressed. A general framework was proposed that allows to obtain optimal allocation considering several aspects, like execution quality and workload, cooperative tasks, human supervision and spatial constraints. The problem is formulated as a Mixed-Integer Linear Programming Problem and an optimal allocation is found; re-allocation is obtained via real-time monitoring of execution parameters and when performance falls below a given threshold  a new plan is computed.
Future work aims at validating the solution in a real scenario and at considering also plan switching cost in the optimization problem.  Furthermore, we plan to define an adaptive procedure to determine the threshold $\delta_t$ for online re-allocation 
according to human preferences.

 \bibliography{biblio}
\end{document}